\documentclass{article}
\usepackage{amsmath,amssymb}
\usepackage{mathtools}
\usepackage{geometry}
\usepackage{natbib}
\title{Anytime Tail Averaging}
\author{Nicolas Le Roux}
\date{\today}
\begin{document}
\maketitle

\begin{abstract}
Tail averaging consists in averaging the last examples in a stream. Common techniques either have a memory requirement which grows with the number of samples to average, are not available at every timestep or do not accomodate growing windows. We propose two techniques with a low constant memory cost that perform tail averaging with access to the average at every time step. We also show how one can improve the accuracy of that average at the cost of increased memory consumption.
\end{abstract}

\section{Introduction}
We assume we receive a stream of samples $x_t$ and our goal is to compute a running average of the last $k_t$ samples
\begin{align}
\label{eq:true}
\bar{x}_t^{\textrm{true}}	&= \frac{1}{k_t} \sum_{i=t-k_t+1}^t x_i \; ,
\end{align}
where $k_t$ can be a constant equal to $k$ or depend on $t$, for instance $k_t = \left \lceil{ct}\right \rceil$ with $c < 1$.

Computing this average exactly has a memory cost that scales linearly with $k_t$. When the $x_t$ are the parameters of a large network, this cost is prohibitive. Reducing that cost is of great practical interest and previous works have proposed solutions with theoretical guarantees, e.g.~\citep{datar2002maintaining}. In practice, however, two techniques are mainly used.

First, one can forgo the ability to get the running average at every timestep $t$. In that case, one decides ahead of time the iteration $T$ at which they wish to have the average and start accumulating the $x_i$'s at $t = T - k_t +1$. If, at $T$, they wish to continue receiving examples, they will not have access to a more recent average until the time $t = T + k_t - 1$~\citep{bach2013non,hazan2014beyond}. This is especially problematic when $k_t$ is large as there will be proportionately few iterations where we will have access to an average. Although this effect can be mitigated by batching samples, having access to the average at every time step recover the orginal method with its associated memory cost.

Another, more common, approach is to replace the exact average with an exponential one:
\begin{align}
\label{eq:exp_k}
	\bar{x}^{\exp}_t	&= \sum_{i=0}^t (1 - \gamma) \gamma^{t-i}x_i \; ,
\end{align}
with $\gamma < 1$. This method has a constant memory cost, making it the method of choice for most running averages, for instance to compute the mean and variance of the activation of each unit in BatchNorm~\citep{ioffe2015batch}. It is unclear however how to extend it to the case $k=ct$\footnote{From now on, we will drop the ceiling notation for simplicity.}.

All approaches must balance two elements: they must have low variance, i.e. average over enough samples, and low staleness, i.e. favor recent examples. One can thus imagine trying to find the method minimizing the variance for a given staleness or to minimize the staleness for a given variance. Unfortunately, we are not aware of any universally accepted measure of staleness that would allow us to rank methods. We shall thus propose methods which achieve the same variance as Eq.~\ref{eq:true} and will assess the impact of their different staleness by avering the iterates of a stochastic linear regression problem.

We first propose an extension of the exponential average to the case $k_t=ct$. The resulting algorithm is simple, memory efficient and a very good approximation to the true running average for low values of $c$. For larger values of $c$, however, its use of old examples degrades its performance. We thus then propose a sliding window version of the running average with constant, adaptive, memory costs and a performance virtually indistinguishable from the true average. These two averaging methods combine the following advantages:
\begin{itemize}
	\item They are available anytime;
	\item They have a computational cost independent of $k_t$;
	\item They work for $k_t=ct$.
\end{itemize}

We coin these methods \textit{anytime tail averages} (ATA) and will evaluate them in the context of optimization where a stochastic method is run with a constant stepsize and the iterates averaged, as done by~\citet{jain2016parallelizing},

\section{Growing exponential average}
We recall here that our goal is to develop methods with the same variance as a running average with a window of size $k_t$, i.e. to maintain a variance $\frac{1}{k_t}$ for all $t$. We start by expanding the exponential average to the growing window case, i.e. to get a variance of $\frac{1}{k_t} = \frac{1}{ct}$ at every step\footnote{The exponential average in Eq.~\ref{eq:exp_k} achieves a variance corresponding to $k = \frac{1 + \gamma}{1 - \gamma}$.}. We thus search for a weighting $\alpha_{i,t}$ of each datapoint $i$ at timestep $t$ such that
\begin{align*}
\sum_{i=1}^t \alpha_{i,t}	&= 1\\
\sum_{i=1}^t \alpha_{i,t}^2	&= \frac{1}{k_t}\\
\alpha_{i,t} &= \gamma_t \alpha_{i, t-1} \quad \forall i < t \; .
\end{align*}
The first constraint ensures that this is an average. The second constraint ensures that the average satisfies the variance requirement. The last constraint ensures that these weights can be obtained using an update of the form
\begin{align}
\label{eq:exp_update}
	\bar{x}^{\exp}_t	&= \gamma_t \bar{x}^{\exp}_{t-1} + (1 - \gamma_t)x_t \; ,
\end{align}

Let us assume that the constraint is satisfied at timestep $t-1$. What is the value of $\gamma_t$ required to achieve the constraint at timestep $t$? Writing the formula for the variance, we get
\begin{align*}
	\frac{\gamma_t^2}{c (t-1)} + (1 - \gamma_t)^2	&= \frac{1}{ct}\\
	\gamma_t	&= \frac{c(t-1)}{1 + c(t-1)}\left(1 \pm \frac{1}{c}\sqrt{\frac{1-c}{t(t-1)}}\right) \; .
\end{align*}
Since our goal is to minimize staleness, we wish to give as much weight as possible to the last example and shall keep the smallest of the two solutions, i.e.
\begin{align}
\label{eq:gamma}
	\gamma_t	&= \frac{c(t-1)}{1 + c(t-1)}\left(1 - \frac{1}{c}\sqrt{\frac{1-c}{t(t-1)}}\right) \; .
\end{align}
Using this sequence $\{\gamma_t\}$, we get that $\frac{k_t}{t}$ converges to $c$ regardless of the initial conditions.

As we shall see later, the exponential average does not perform as well as the window average in the case where $k$ is fixed. Thus, one can wonder if one can do better in the case $k_t = ct$ than using Eq.~\ref{eq:exp_update} with $\gamma_t$ defined as in Eq.~\ref{eq:gamma}? Exponential averages use all examples since the beginning. In settings where there is a fast evolution of the average followed by a more stationary process, as is the case in optimization, a sliding window strategy might be preferable. We explore such a strategy in the next section.

\section{Anytime window average}
We now propose a method, \textbf{anytime window average} (AWA) which weighs all used examples almost equally in order to remove the dependency on very old examples. As we have no knowledge of such a method existing even in the case of constant $k_t = k$, we start by describing this setting before moving to $k_t = ct$.

Our method will use multiple accumulators of different recency. When computing the average, priority will be given to the most recent accumulators but older ones will also be used to reach the desired variance.

Let us start with some notations:
\begin{itemize}
	\item $\bar{x}^j_t$ will be the average vector in accumulator $j$ at time $t$;
	\item $N^j_t$ will be the number of elements in accumulator $j$ at time t;
	\item The accumulators will be in increasing order of recency, the oldest accumulator having index $0$.
\end{itemize}

\subsection{$k_t = k$ with two accumulators}

\begin{figure}
	\includegraphics[width=\textwidth]{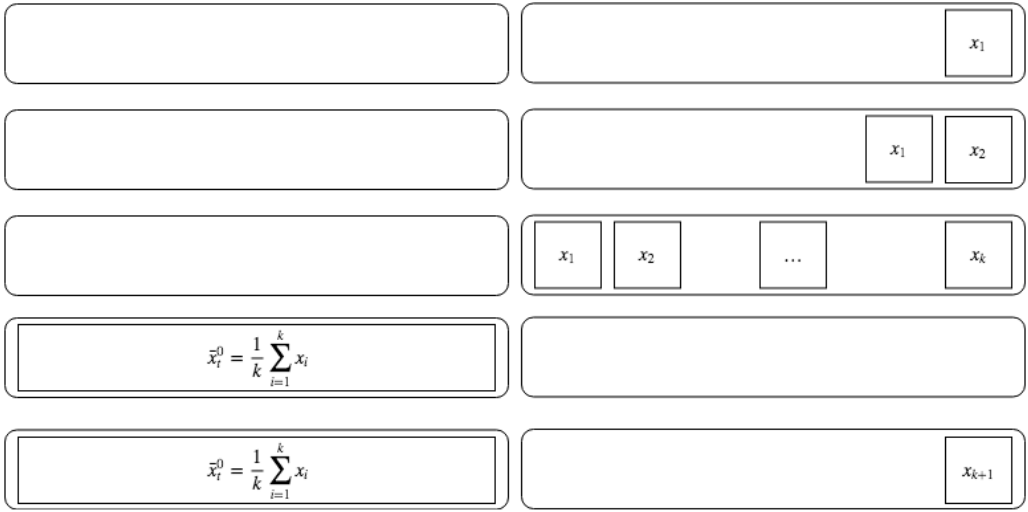}
	\caption{Representation of AWA with two accumulators. Examples enter the second accumulator which is flushed to the first accumulator and reset when full.\label{fig:awa}}
\end{figure}

In its simplest form, AWA uses only two accumulators. As a sample $x_t$ arrives, we update the second accumulator and leave the first one untouched:
\begin{align*}
	N^1_t		&\gets N^1_{t-1} + 1\\
	\bar{x}^1_t	&\gets \bar{x}^1_{t-1} + \frac{1}{N^1_t}(x_t - \bar{x}^1_{t-1})\\
	\bar{x}^0_t	&\gets \bar{x}^0_{t-1}\; .
\end{align*}

When the second accumulator contains $k$ samples, i.e. $N^1_t = k$, we copy the average vector $\bar{x}^1_t$ to the first accumulator and reset the second one:
\begin{align*}
	\bar{x}^0_t	&\gets \bar{x}^1_t\\
	\bar{x}^1_t	&\gets 0\\
	N^1_t	&\gets 0 \; .
\end{align*}

We then continue accumulating future samples in $\bar{x}^1_t$. Figure~\ref{fig:awa} shows an example of the two accumlators for the first $k+1$ examples.

As $\bar{x}^1_t$ is the average of fewer than $k_t$ samples, we will use $\bar{x}^0_t$ to complement it. To find the optimal weight $\gamma$ between the two accumulators, we will maximize the weight given to the most recent examples while maintaining the variance of $1/k$. Since $\bar{x}^0_t$ has a variance of $1/k$ and $\bar{x}^1_t$ has a variance of $1/N^1_t$, we get\footnote{We assume $t > k$ here.}
\begin{align*}
\gamma^*&= \max \gamma \textrm{ s.t. } \frac{\gamma^2}{N^1_t} + \frac{(1 - \gamma)^2}{k} = \frac{1}{k}\\
		&= \frac{2N^1_t}{N^1_t + k} \; .
\end{align*}
Our running average is thus:
\begin{align}
\label{eq:average_two}
	\bar{x}_t^\textrm{awa}	&= \bar{x}^1_t + \frac{k - N^1_t}{N^1_t + k}(\bar{x}^0_t - \bar{x}^1_t) \; .
\end{align}
We can view $\bar{x}_t^\textrm{awa}$ as the standard incomplete window average $\bar{x}^1_t$ with a correction term to ensure a constant variance. As we accumulate samples in $\bar{x}^1_t$, the weight of the correction term decreases. Whenever $N^1_t = k$, we recover the classic, non-anytime, tail average.

We now turn to the case $k_t=ct$ with $c < 1$.

\subsection{$k_t=ct$ with two accumulators}
The case $k_t=ct$ is similar to the previous case with the exception that the accumulator gets reset when $N^1_t \geq ct$. The computation of the optimal weight $\gamma$ of the combination between the averages of the accumulators changes slightly to
\begin{align}
\gamma^*&= \max \gamma \textrm{ s.t. } \frac{\gamma^2}{N^1_t} + \frac{(1 - \gamma)^2}{N^0_t} = \frac{1}{k}\nonumber\\
&= \frac{N^1_t + N^0_tN^1_t\sqrt{\frac{1}{N^0_tct} + \frac{1}{N^1_tct} - \frac{1}{N^0_tN^1_t}}}{N^1_t + N^0_t} \; .\label{eq:optimal_gamma}
\end{align}
Our running average is thus:
\begin{align}
\label{eq:average_two_ct}
\bar{x}_t^\textrm{awa}	&= \bar{x}^1_t + \frac{N^0_t - N^0_tN^1_t\sqrt{\frac{1}{N^0_tct} + \frac{1}{N^1_tct} - \frac{1}{N^0_tN^1_t}}}{N^1_t + N^0_t}(\bar{x}^0_t - \bar{x}^1_t) \; .
\end{align}
Once again, we see that the correction term in Eq.~\ref{eq:average_two_ct} decreases as $N^1_t$ until it completely vanishes when $N^1_t = ct$.

Without going to the extreme of the exponential average, we see that AWA can use  $ct + N^0_t$ samples, an increase of $N^0_t$ compared to the true average. We shall now see how adding accumulators will help reduce the maximum staleness of the examples used.

\subsection{$k_t=k$ with an arbitrary number of accumulators}
\label{sec:multiple_constant}
In this section, we now assume that we have access to $z+1$ accumulators. For simplicity, we shall assume that $k$ is a multiple of $z$.

As before, whenever a sample $x_i$ arrives, it is placed in the last accumulator:
\begin{align*}
N^z_t		&\gets N^z_{t-1} + 1 \\
\bar{x}^z_t	&\gets \bar{x}^z_{t-1} + \frac{1}{N^z}(x_t - \bar{x}^z_{t-1})\; .
\end{align*}

When the number of samples into the last accumulator reaches $\frac{k}{z}$, we move all the average vectors $\bar{x}^j_t$ to the accumulator $j-1$ and we reset the last one:
\begin{align*}
\bar{x}^{j-1}_t	&\gets \bar{x}^j_t \quad \forall j > 0\\
N^{j-1}_t &\gets N^j_t  \quad \forall j > 0\\
\bar{x}^z_t	&\gets 0\\
N^z_t		&\gets 0 \; .
\end{align*}

We now have to find the optimal weight to give to each average. There are multiple possible strategies: for instance, one might want to maximize the weight of the most recent accumulator $\bar{x}^z_t$ or to minimize the weight of the oldest accumulator $\bar{x}^0_t$. We will take here the latter approach with the reasoning that, in optimization, it is often more important to forget the oldest iterates than to use the freshest ones.

The weighting for the averages $\bar{x}^1_t, \ldots, \bar{x}^z_t$ leading to the minimum variance is proportional to the number of samples in each accumulator. Finding the optimal weighting of each accumulator is thus equivalent to finding the relative weighting of the last accumulator compared to the weighting of all the other ones. Modifying Eq.~\ref{eq:optimal_gamma} to have one accumulator with $N^0_t$ samples and the other one with $N^{-0}_t = \sum_{i=1}^z N^i_t$ samples, we get that the optimal weight $\gamma^i_t$ of each average $\bar{x}^i_t$ is thus
\begin{align*}
\gamma^0_t	&= \frac{N^0_t \left(1 - N^{-0}_t\sqrt{\frac{1}{N^0_tk} + \frac{1}{N^{-0}_tk} - \frac{1}{N^0_tN^{-0}_t}}\right)}{N^0_t + N^{-0}_t}\\
\gamma^i_t 	&= \frac{N^i_t \left(1 + N^{0}_t\sqrt{\frac{1}{N^0_tk} + \frac{1}{N^{-0}_tk} - \frac{1}{N^0_tN^{-0}_t}}\right)}{N^0_t + N^{-0}_t} \qquad \forall i > 0 \; ,
\end{align*}
and the AWA is
\begin{align}
\bar{x}_t	&= \frac{1}{\sum_{i=1}^z N^i_t}\sum_{i=1}^z N^i_t \bar{x}^i_t\nonumber\\
&\qquad  + \frac{N^0_t \left(1 - N^{-0}_t\sqrt{\frac{1}{N^0_tk} + \frac{1}{N^{-0}_tk} - \frac{1}{N^0_tN^{-0}_t}}\right)}{N^0_t + N^{-0}_t}\left(\bar{x}^0_t - \frac{1}{\sum_{i=1}^z N^i_t}\sum_{i=1}^z N^i_t \bar{x}^i_t\right) \; .
\end{align}

Once again, we find that the correction term goes is zero whenever $N^{-0}_t = k$.

\subsection{$k_t=ct$ with an arbitrary number of accumulators}
The case $k_t = ct$ is similar to the constant $k$ case with the difference that the last accumulator is not reset when $N^z_t$ reaches $\frac{k}{z}$ but rather when the sum of examples into all accumulators but the oldest one reaches $ct$, i.e. $\sum_{i=1}N^i_t \geq ct$. At that point, we apply the same update equations as in Section~\ref{sec:multiple_constant}, replacing $k$ with $ct$:
\begin{align*}
\gamma^0_t	&= \frac{N^0_t \left(1 - N^{-0}_t\sqrt{\frac{1}{N^0_tct} + \frac{1}{N^{-0}_tct} - \frac{1}{N^0_tN^{-0}_t}}\right)}{N^0_t + N^{-0}_t}\\
\gamma^i_t 	&= \frac{N^i_t \left(1 + N^{0}_t\sqrt{\frac{1}{N^0_tct} + \frac{1}{N^{-0}_tct} - \frac{1}{N^0_tN^{-0}_t}}\right)}{N^0_t + N^{-0}_t} \qquad \forall i > 0 \; ,
\end{align*}
and
\begin{align}
    \bar{x}_t	&= \frac{1}{\sum_{i=1}^z N^i_t}\sum_{i=1}^z N^i_t \bar{x}^i_t\nonumber\\
    &\qquad  + \frac{N^0_t \left(1 - N^{-0}_t\sqrt{\frac{1}{N^0_tct} + \frac{1}{N^{-0}_tct} - \frac{1}{N^0_tN^{-0}_t}}\right)}{N^0_t + N^{-0}_t}\left(\bar{x}^0_t - \frac{1}{\sum_{i=1}^z N^i_t}\sum_{i=1}^z N^i_t \bar{x}^i_t\right) \; .
\end{align}

\section{Experiments}
We now test the empirical performance of our various averages. Following the work of~\citet{jain2018parallelizing}, we will study the case of stochastic linear regression where we minimize the loss
\begin{align*}
	\ell(w) = E_{x,y} (x^\top w - y)^2 \; ,
\end{align*}
with $x \sim \mathcal{N}(0, H)$, $H$ a $50\times 50$ diagonal matrix whose values are $H_{ii} = \frac{1}{i}$ and $y \sim \mathcal{N}(x^\top w^\ast, \epsilon)$ with $\epsilon^2 = 0.01$. We also used a batch size of 11 to match their experiments. We ran the optimization for 1000 batches and averaged the results over 100 runs. All plots display the excess error on a log-log scale.

When $k_t$ si constant, i.e. $k_t = k$, the following averagers are used:
\begin{itemize}
	\item \textit{expk}: the exponential average with $\gamma = \frac{k-1}{k+1}$;\\
	\item \textit{awak}: the anytime window average with two accumulators;\\
	\item \textit{truek}: the exact moving average.
\end{itemize}

When $k_t = ct$, the following averagers are used:
\begin{itemize}
	\item \textit{raw}: no averaging occurs until the step $t = T(1-c)$ with $T$ the maximum number of steps. This is the standard way to do tail averaging;\\
	\item \textit{exp}: the growing exponential average;\\
	\item \textit{awa / awa3}: the anytime window average with two / three accumulators;\\
	\item \textit{true}: the exact moving average, with a memory cost that grows with $k_t$.
\end{itemize}

\begin{figure}[ht]
	\includegraphics[width=.5\textwidth]{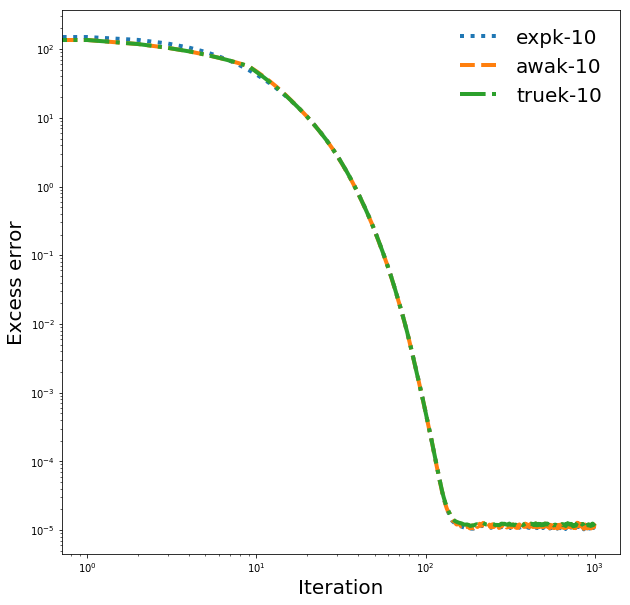}
	\includegraphics[width=.5\textwidth]{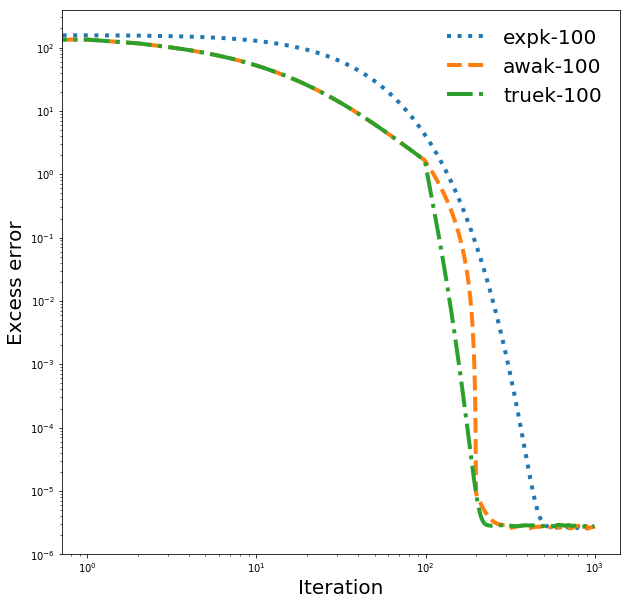}
	\caption{Comparison between the exponential average (\textit{expk}), the anytime window estimator with two accumulators (\textit{awa}) and the true window average (\textit{truek}) for a constant variance $1/k$. Results are given for $k=10$ (\textbf{left}) and $k=100$ (\textbf{right}). We see that, as $k$ grows, the performance of the exponential average degrades faster than that of the anytime window average.
		\label{fig:results_k}}
\end{figure}

The results for constant $k_t=k$ are shown in Figure~\ref{fig:results_k}. While all methods perform the same for small values of $k$, the exponential average degrades faster when $k$ increases. One can conclude that, when $x$ evolves quickly, as is the case at the beginning of optimization, the use of old examples, even with a small weights, heavily penalizes the average.

\begin{figure}[ht]
	\includegraphics[width=.5\textwidth]{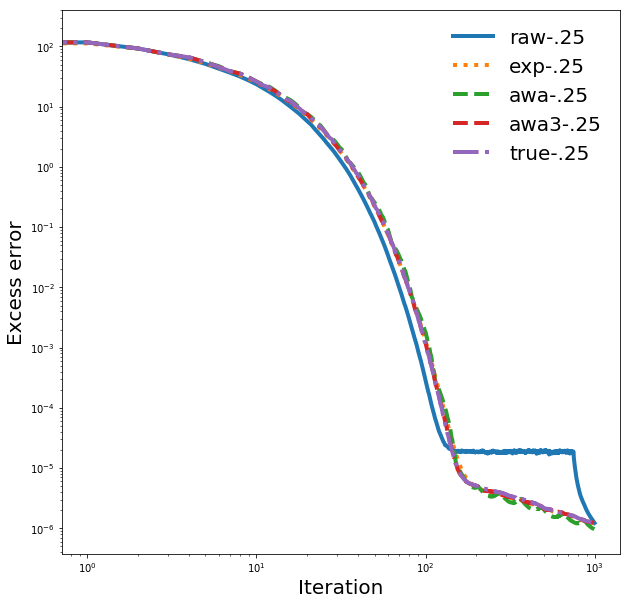}
	\includegraphics[width=.5\textwidth]{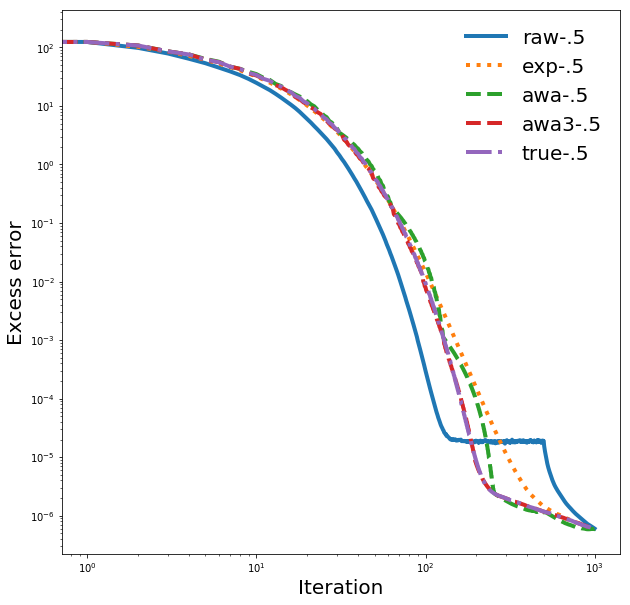}
	\caption{Comparison between the excess loss of the standard tail average (\textit{raw}), the growing exponential average (\textit{exp}), the anytime window estimator with two (\textit{awa}) or three (\textit{awa3}) accumulators and the true window average (\textit{true}) for a variance $1/ct$. Results are given for $c=.25$ (\textbf{left}) and $c=.5$ (\textbf{right}). We see that, although all proposed estimators closely match the true average for $c=.25$, only \textit{awa3} performs as well as the true average for $c=.5$.
	\label{fig:results}}
\end{figure}

The results for $k_t=ct$ are shown in Figure~\ref{fig:results_k} and paint a similar story. For $c=.25$, all methods except \textit{raw} are indistinguishable. This is to the advantage of the growing exponential average which has the lowest memory cost of all methods. For $c=.5$, however, staleness is more of an issue since many iterates are before the optimization reached the noise ball. In that setting, \textit{exp} performs significantly worse than \textit{true}. \textit{awa} with two accumulators also has lower performance but using only three accumulators achieves the exact same rate as the true window average.

\section{Conclusion}
We presented here two Anytime Tail Averages, a set of techniques to obtain a tail average at every timestep with a constant memory cost. We showed how one could use either a growing exponential average, for the lowest memory cost, or an anytime window average with an adjustable memory cost for higher accuracy. We showed how this could be used when the number of elements in the average increases over time, as well as how to adjust the tradeoff between the accuracy of the average and the memory consumption.

Although we only performed experiments in the standard optimization setting, we believe these methods could be of interest when tracking the average over two phases: a quickly changing one followed by a more stable one. For instance, BatchNorm~\citep{ioffe2015batch} tracks the mean and variance of the activation of each unit over time. One could imagine that, as the optimization stabilizes, these quantities should be estimated over longer time periods, which is now possible with the growing exponential average.

\subsection*{Acknowledgments}
The author would like to thank Thijs Vogels for pointing out that the case $k_t=k$ with multiple accumulators could not be immediately derived from the case $k_t=ct$. They would also like to thank Pierre-Antoine Manzagol, S\'ebastien Arnold, Ioannis Mitliagkas, Bart van Merri\"enboer, Sergey Ioffe and Satyen Kale for helpful discussions.

\bibliography{../../latex/full.bib}

\begin{thebibliography}{6}
\providecommand{\natexlab}[1]{#1}
\providecommand{\url}[1]{\texttt{#1}}
\expandafter\ifx\csname urlstyle\endcsname\relax
  \providecommand{\doi}[1]{doi: #1}\else
  \providecommand{\doi}{doi: \begingroup \urlstyle{rm}\Url}\fi

\bibitem[Bach and Moulines(2013)]{bach2013non}
Francis Bach and Eric Moulines.
\newblock Non-strongly-convex smooth stochastic approximation with convergence
  rate o (1/n).
\newblock In \emph{Advances in Neural Information Processing Systems}, pages
  773--781, 2013.

\bibitem[Datar et~al.(2002)Datar, Gionis, Indyk, and
  Motwani]{datar2002maintaining}
Mayur Datar, Aristides Gionis, Piotr Indyk, and Rajeev Motwani.
\newblock Maintaining stream statistics over sliding windows.
\newblock \emph{SIAM journal on computing}, 31\penalty0 (6):\penalty0
  1794--1813, 2002.

\bibitem[Hazan and Kale(2014)]{hazan2014beyond}
Elad Hazan and Satyen Kale.
\newblock Beyond the regret minimization barrier: optimal algorithms for
  stochastic strongly-convex optimization.
\newblock \emph{The Journal of Machine Learning Research}, 15\penalty0
  (1):\penalty0 2489--2512, 2014.

\bibitem[Ioffe and Szegedy(2015)]{ioffe2015batch}
Sergey Ioffe and Christian Szegedy.
\newblock Batch normalization: Accelerating deep network training by reducing
  internal covariate shift.
\newblock In \emph{International Conference on Machine Learning}, pages
  448--456, 2015.

\bibitem[Jain et~al.(2016)Jain, Kakade, Kidambi, Netrapalli, and
  Sidford]{jain2016parallelizing}
Prateek Jain, Sham~M Kakade, Rahul Kidambi, Praneeth Netrapalli, and Aaron
  Sidford.
\newblock Parallelizing stochastic approximation through mini-batching and
  tail-averaging.
\newblock \emph{arXiv preprint arXiv:1610.03774}, 2016.

\bibitem[Jain et~al.(2018)Jain, Kakade, Kidambi, Netrapalli, and
  Sidford]{jain2018parallelizing}
Prateek Jain, Sham~M Kakade, Rahul Kidambi, Praneeth Netrapalli, and Aaron
  Sidford.
\newblock Parallelizing stochastic gradient descent for least squares
  regression: Mini-batching, averaging, and model misspecification.
\newblock \emph{Journal of Machine Learning Research}, 18\penalty0
  (223):\penalty0 1--42, 2018.

\end{thebibliography}
\bibliographystyle{plainnat}

\end{document}